\documentclass[letterpaper]{article} 
\usepackage{aaai2026}  
\usepackage{times}  
\usepackage{helvet}  
\usepackage{courier}  
\usepackage[hyphens]{url}  
\usepackage{graphicx} 
\usepackage{natbib}  
\usepackage{caption} 
\usepackage{algorithm}
\usepackage{algorithmic}
\usepackage{comment}
\usepackage{multirow}
\usepackage{amsmath}
\usepackage{amssymb}
\usepackage{bm}
\usepackage{threeparttable}

\usepackage{newfloat}
\usepackage{listings}
\DeclareCaptionStyle{ruled}{labelfont=normalfont,labelsep=colon,strut=off} 
\floatstyle{ruled}
\newfloat{listing}{tb}{lst}{}
\floatname{listing}{Listing}
\title{ParaDySe: A Parallel Strategy Switching Framework for Dynamic Sequences in Transformer-based Large Language Models}

\author{
    Zhixin Ou, 
    Peng Liang\footnote{Peng Liang and Linbo Qiao are the corresponding authors.},
    Jianchen Han,
    Baihui Liu,
    Linbo Qiao\textsuperscript{$*$}
}
\affiliations{
    College of Computer Science and Technology, National University of Defense Technology\\
    \{ouzhixin16, peng\_leung, hjc, lbh, qiao.linbo\}@nudt.edu.cn
}

\usepackage{bibentry}

\begin{document}

\maketitle

\begin{abstract}
Dynamic sequences with varying lengths have been widely used in the training of Transformer-based large language models (LLMs).
However, current training frameworks adopt a pre-defined static parallel strategy for these sequences, causing neither communication-parallelization cancellation on short sequences nor out-of-memory on long sequences.
To mitigate these issues, we propose ParaDySe, a novel adaptive Parallel strategy switching framework for Dynamic Sequences.
ParaDySe enables on-the-fly optimal strategy adoption according to the immediate input sequence.
It first implements the modular function libraries for parallel strategies with unified tensor layout specifications,
and then builds sequence-aware memory and time cost models with hybrid methods. 
Guided by cost models, ParaDySe selects optimal layer-wise strategies for dynamic sequences via an efficient heuristic algorithm.
By integrating these techniques together, ParaDySe achieves seamless hot-switching of optimal strategies through its well-designed function libraries.
We compare ParaDySe with baselines on representative LLMs under datasets with sequence lengths up to 624K. 
Experimental results indicate that ParaDySe addresses OOM and CPC bottlenecks in LLM training by systematically integrating long-sequence optimizations with existing frameworks.
\end{abstract}

\begin{links}
    \link{Code}{https://github.com/Carrie-ou/ParaDySe}
\end{links}

\section{Introduction}

In recent years, Transformer-based large language models (LLMs) have demonstrated remarkable performance not only in text-related tasks but also in cross-domain applications like genomic tasks due to their exceptional parallel computing capability and scalability. 
Models like GPT~\cite{openai2024gpt4technicalreport} leverage self-attention to capture long-distance semantic relationships, with context windows expanding from 512 to 128K tokens, thereby improving performance in long-context tasks, including document understanding and programming assistance.
Growing sequence lengths exponentially increase the computational and memory complexity $O(n^2)$, creating significant challenges for LLM training.

Parallelism strategies partition LLM states and intermediate results into distributed devices, providing additional computational and memory resources for scaled model training. Recent researches improve training efficiency by partitioning in the sequence dimension, such as Sequence Parallelism~\cite{RSA} and Ulysses~\cite{Ulysses}, which have shown great improvements in training throughput. However, when processing documents that exceed 128K tokens, the extremely long sequences often lead to \textit{out-of-memory} (OOM) failures during model training.
To achieve better memory savings, researchers establish fine-grained optimized sequence parallelisms, such as METP~\cite{METP}. It can significantly decrease the memory consumption of devices, making it suitable for training extremely long sequences. However, METP raises communication overhead as a trade-off to training efficiency, thereby negating performance gained by distributed parallelism. The \textit{communication-parallelization cancellation} (CPC) becomes more severe when memory-saving parallel strategies are applied to short sequences, which bring more frequent communications during LLM training.
To conclude, efficient strategies perform optimally for short sequences, while memory-saving strategies are suited for long sequences.

Existing LLM training frameworks, such as Megatron-LM~\cite{shoeybi2020megatronlmtrainingmultibillionparameter, 10.1145/3458817.3476209,MegatronTP}, typically offer multiple parallel strategies for options, including tensor parallelism (TP), sequence parallelism (SP), etc~\cite{tang2025koala,liang2023surveyAP}. 
These frameworks significantly enhance the capabilities of large-model training. Strategic parallelism selection enables optimal efficiency-memory trade-off for given workloads. By selecting appropriate parallel strategies for the training dataset, they can achieve a balanced trade-off between training efficiency and memory consumption. 
However, ideally assuming the training workloads are static across different samples, these frameworks mostly employ a fixed parallel strategy for all Transformer layers throughout the entire training process. 
Therefore, they fail to adapt to real-world input sequences that range from queries consisting of several tokens to those comprising millions of tokens, remaining unable to resolve OOM failures or CPC issues.

As a promising alternative, HotSPa~\cite{HotSpa} pioneers hot-switching at the mini-batch level for different sequence lengths across parallel strategies, integrating unified graph compilation and communication-aware scheduling.
However, its coarse-grained switching mechanism, while supporting sequences up to 32K tokens, exhibits incompatibility with modern memory-efficient parallel strategies.
This calls for a training framework that incorporates both novel and conventional parallelism to support extremely long sequences exceeding 300K tokens, adaptively selects parallel strategies for dynamic sequences, and achieves fine-grained switching. 

To address it, this paper presents ParaDySe, a \textbf{Para}llel-strategy switching approach for \textbf{Dy}namic \textbf{Se}quences in Transformer models. ParaDySe enables layer-wise adaptive parallel training based on real-time sequence lengths, achieving seamless strategy switching without tensor redistribution or communication synchronization. 
Specifically, our contributions can be summarized as follows:

\begin{itemize}
    \item We propose ParaDySe, a novel framework enabling adaptive parallel strategy switching for dynamic sequences, which achieves sequence length support up to 624K tokens while significantly improving training efficiency.
    \item Through the design of modular function libraries based on tensor layout specifications, ParaDySe eliminates tensor redistribution overhead, thereby enabling seamless strategy switching.
    \item Through sequence-aware cost models for time and memory, ParaDySe achieves on-the-fly optimal layer-wise strategy selection adapted to dynamic sequences, with a balanced trade-off between training efficiency and memory consumption.
\end{itemize}

\section{Related Work}
\subsection{LLM Training Frameworks}
Transformer-based LLMs have two core operations: the Multi-Head Attention (MHA) and the Feed-Forward Network (FFN) operation.
As LLMs demand increasing computational resources, various automatic parallel training frameworks have been proposed to address challenges in efficiency and memory consumption. \textbf{DeepSpeed} introduces the ZeRO family~\cite{ZeRO-Offload,ZeRO-Infinity} for fine-grained partitioning of optimizer states, gradients, and parameters, significantly reducing memory usage. \textbf{Megatron-LM} focuses on scalable training via tensor parallelism (TP)~\cite{10.1145/3458817.3476209}, and extends it with sequence parallelism (SP) and context parallelism (CP)~\cite{MegatronTP} to support longer sequences.\textbf{Colossal-AI} incorporates Ring Self-Attention (RSA)~\cite{RSA} to parallelize MHA computation. Recent work has also explored automated strategy generation. \textbf{AutoPipe}~\cite{liu2022autopipe} introduces a heuristic-based planner that partitions transformer blocks into sub-layer granularity to balance pipeline stages. \textbf{Merak}~\cite{lai2023merak} further generalizes this idea by automatically composing 3D parallelism through a proxy-graph abstraction. While these frameworks achieve impressive performance, most adopt static parallel strategies fixed at training initialization. \textbf{HotSPa}~\cite{HotSpa} addresses this limitation via mini-batch-level hot-switching by pre-defined checkpoints and strategy sets, but restricts support to conventional parallel paradigms. Extending this line of dynamically switching, our ParaDySe framework introduces a more fine-grained framework at the Transformer layer level, which achieves sequence-aware adaptive strategy hot-switching, along with a more flexible functional programming approach.

\subsection{Parallel Methods on 1D Device Grids}
The device grid describes the topological structure of a computing cluster. One-dimensional (1D) device grids represent the linear arrangement of devices, typical in single-node multi-accelerator configurations. 
Several operator-level parallel methods have been developed for 1D device grids. \textbf{Megatron-LM TP+SP} employs AllGather and ReduceScatter communications along the sequence dimension in both MHA and FFN layers. Its CP extension further enhances parallelism through sequence-dimension partitioning. \textbf{DeepSpeed Ulysses}~\cite{Ulysses} distributes attention heads via All-to-All communication, enabling independent processing of head subsets across devices. \textbf{Colossal-AI SP}~\cite{RSA} introduces ring-based communication patterns for MHA optimization, though its activation memory remains quadratic with sequence length. \textbf{METP}~\cite{METP} advances memory optimization through fine-grained partitioning and asynchronous ring-based execution, featuring a two-level loop structure for MHA that enables computation-communication overlap. However, existing long-sequence training methods lack strategy composability and runtime adaptability. Our work extends these methods by enhancing modularity and enabling dynamic control within 1D device grid constraints.

\section{Problem Formulation} 
\begin{table}
    \centering
    \begin{tabular}{cc}
    \hline
    Notation  &  Definition  \\ \hline
    $b\in \mathbb{Z}^+$ & Batch size \\
    $s\in \mathbb{Z}^+$ & Sequence length \\
    $h\in \mathbb{Z}^+$ & Hidden dimension size \\
    $n\in \mathbb{Z}^+$ & Number of attention heads \\
    $p\in \mathbb{Z}^+$ & Parallelism degree \\ 
    $l \in [1,L]$  &  Layer index of $L$-layer Transformer \\
    $\pi \in \mathcal{P}$  &  Parallel strategy within strategy set $\mathcal{P}$\\
    $q \in \mathcal{Q}$  &  Transformer operation, $\mathcal{Q}=\{\text{MHA},\text{FFN}\}$ \\
    $\bm{X} \in \mathcal{X}$  &  Layer input (e.g., $\bm{X_{\text{MHA}}}$) \\
    $\bm{W} \in \mathcal{W}$  &  Layer parameter (e.g., $\bm{W_{\text{qkv}}}$) \\
    $\bm{Y} \in \mathcal{Y}$  &  Layer output (e.g., $\bm{O},\bm{Z}$) \\
    $\mathcal{M}_\pi(\bm{X})$  &  Memory cost model under $\pi$ \\
    $\mathcal{T}_\pi(\bm{X})$  &  Time cost model under $\pi$ \\
    $f_{\pi,q}^{l} \in F$  &  Transformer function under ($\pi,q,l$) \\
    \hline
    \end{tabular}
    \caption{Notations and definitions.}
    \label{tab:notation}
\end{table}

\begin{figure*}[t]
    \centering
    \includegraphics[width=\linewidth]{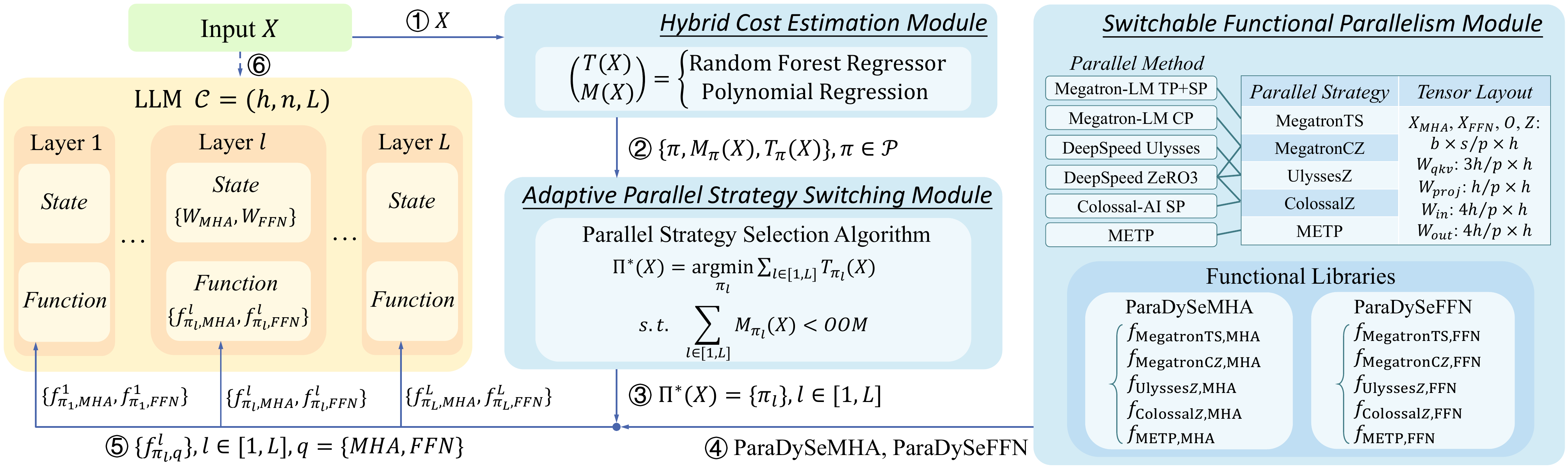}
    \caption{Overview of ParaDySe Framework for Dynamic sequences with varying lengths. These three core Modules (namely Hybrid Cost Estimation Module,  Switchable Functional Parallelism Module,  and Adaptive Parallel Strategy Switching Module) are collaborative adaptively to achieve almost optimal parallel strategy according to the input sequence. The details of these modules are presented in the following subsection.}
    \label{fig:overview}
\end{figure*}

The symbolic notations used in this paper are defined in Table \ref{tab:notation}.
Transformer architecture has two core operations ($q\in \mathcal{Q}$), the Multi-Head Attention (MHA) and the Feed-Forward Network (FFN) operation, whose computational process can be formally expressed as Equations (\ref{eq: qkv}-\ref{eq: MHAoutput}) and Equation (\ref{eq: ffn}) respectively.
\begin{align}
    \bm{Q},\bm{K},\bm{V} &= \bm{X_{\text{MHA}} W_{\text{qkv}}} \label{eq: qkv} \\
    \bm{A} &= \text{Softmax}(\frac{\bm{QK^\text{T}}}{\sqrt{d}})\bm{V} \label{eq: attention} \\
    \bm{O} &= \bm{AW}_{\text{proj}} \label{eq: MHAoutput} \\
  \bm{Z} &= \text{GELU}(\bm{X_{\text{FFN}}W}_{\text{in}})\bm{W}_{\text{out}}\label{eq: ffn}
\end{align}

Let $\mathcal{C} = (h, n, L)$ be the configuration tuple that determines LLM type, as a simplification for Transformer-based LLM. Notably, we treat $\mathcal{C}$ as a fixed hyperparameter for methodology development, and only vary it in experimental evaluation.
Given Transformer layer index $l$, parallel strategy $\pi$, and operation $q$, there exists $f_{\pi,q}^{l} \in F$ that implements the corresponding function, where $F: \mathcal{X} \times \mathcal{W} \rightarrow \mathcal{Y}$. 
For each parallel strategy $\pi$ under fixed configuration $\mathcal{C}$, there exists layer-wise cost models $\mathcal{M}_\pi(\bm{X})$ and $\mathcal{T}_\pi(\bm{X})$, corresponding to the input $\bm{X}$ with sequence length $s$.

ParaDySe aims to determine the optimal parallel strategy lists $\Pi^*=\{\pi_1^*,\cdots,\pi_L^*\}$, where $\pi_l^*$ denotes the parallel strategy at layer $l$ that satisfies global optimization, for dynamically input sequence $\bm{X}$ by solving a constrained cost minimization problem:
\begin{align}
   & \Pi^*(\bm{X}) = \mathop{\text{argmin}}_{\pi_l \in \mathcal{P}} \mathop{\Sigma}_{l \in [1,L]}  \mathcal{T}_{\pi_l}(\bm{X}) \\
& \text{s.t.} \quad \mathop{\Sigma}_{l \in [1,L]}\mathcal{M}_{\pi_l}(\bm{X}) < OOM
\end{align}

\section{ParaDySe Framework}

\subsection{Overview}

This paper proposes the ParaDySe framework, a \textbf{Para}llel-Strategy framework for \textbf{Dy}namic \textbf{Se}quences lengths in Transformer models. The ParaDySe framework consists of three core modules, the Switchable Functional Parallelism Module, the Hybrid Cost Estimation Module, and the Adaptive Parallel Strategy Switching Module, as shown in Figure~\ref{fig:overview}.

\begin{table*}
    \centering
    \begin{tabular}{c|cccccccc}
        \hline
         \textit{Parallel Method} & \textit{$\bm{X}_{\text{MHA}}$, $\bm{O}$, $\bm{X}_{\text{FFN}}$, $\bm{Z}$} & $\bm{W}_{\text{qkv}}$ & $\bm{W}_{\text{proj}}$ & $\bm{W}_{\text{in}}$ & $\bm{W}_{\text{out}}$ \\ \hline
         Megatron-LM TP &$b\times s\times h$  &$h\times 3h/p$  &$h/p\times h$  &$h\times 4h/p$  &$4h/p\times h$  \\
         Megatron-LM TP+SP &$b\times s/p\times h$  &$h\times 3h/p$  &$h/p\times h$ &$h\times 4h/p$ &$4h/p\times h$  \\
         Megatron-LM CP &$b\times s/p\times h$  &$h\times 3h$  &$h\times h$ &$h\times 4h$  &$4h\times h$ \\
         DeepSpeed Ulysses &$b\times s/p\times h$  &$h\times 3h$  &$h\times h$ &$h\times 4h$ &$4h\times h$  \\
         DeepSpeed ZeRO3 &$b/p\times s\times h$  &$(h/p\times 3h)^{\text{T}}$  &$h/p\times h$  &$(h/p\times 4h)^{\text{T}}$  &$4h/p\times h$ \\
         Colossal-AI SP &$b\times s/p\times h$  &$h\times 3h$  &$h\times h$ &$h\times 4h$  &$4h\times h$ \\
         METP &$b\times s/p\times h$  &$h\times 3h/p$  &$h/p\times h$  &$h\times 4h/p$  &$4h/p\times h$ \\

         \textbf{Specification} &$b\times s/p\times h$  &$(3h/p\times h)^{\text{T}}$  &$h/p\times h$ &$(4h/p\times h)^{\text{T}}$ &$4h/p\times h$  \\
         \hline
    \end{tabular}
    \caption{The tensor layout analysis on parallel methods, and tensor layout specification of parallel strategies.}
    \label{tab:layout}
\end{table*}

\begin{algorithm}[tbp]
\caption{Parallel Strategy Selection Algorithm}
\label{alg:select}
    \begin{algorithmic}[1]
    \REQUIRE $b$: batch size; $s$: sequence length; $\mathcal{C}=(h,n,L)$: LLM hyperparameters; $D$: Dictionary with keys as $(b, s)$ and values as the optimal strategy lists $\Pi^{(b, s)}=\{\pi_1,\cdots,\pi_L\}$; $P$: A list of parallel strategies sorted in ascending order by latency and memory usage.
    \ENSURE $\Pi^*=\{\pi_1^*,\cdots,\pi_L^*\}$, where $\pi_l^*$ denotes the optimal parallel strategy at layer $l$ that satisfies global optimization.
    \STATE pop\_useless(strategy\_sorted)
    \IF{$(b, s)$ in $D$}
        \RETURN $D[(b,s)]$
    \ENDIF
    \STATE strategy\_options $\gets$ [ ]
    \FOR{$i = 0$ to $\text{len}(P) - 1$}
        \STATE strategies $\gets$ [$P[i]$] $\times$ $L$
        \IF{$i == 0$ and not OOM(strategies)}
            \STATE $\Pi^*$ $\gets$ strategies
            \STATE break
        \ENDIF
        \IF{not OOM(strategies)}
            \STATE append strategies to strategy\_options
        \ELSE
            \FOR {$k = i+1$ to $\text{len}(P) - 1$}
                \FOR{$l = 0$ to $L-1$}
                    \STATE pop the first element from strategies
                    \STATE append $P[k]$ to the end of strategies
                    \IF{not OOM(strategies)}
                        \STATE append strategies to strategy\_options
                    \ENDIF
                \ENDFOR
            \ENDFOR
        \ENDIF
    \ENDFOR
    \IF{strategy\_options is not empty}
        \STATE $\Pi^*$ $\gets$ least time cost strategies in strategy\_options
    \ELSE
        \STATE $\Pi^*$ $\gets$ least memory cost strategies $\times$ $L$
    \ENDIF
    \STATE $D[(b, s)]$ $\gets$ $\Pi^*$
    \STATE \RETURN $\Pi^*$
\end{algorithmic}
\end{algorithm}

\subsection{Switchable Functional Parallelism Module} 

In distributed deep learning, the tensor layout describes the specific distribution pattern of tensors across the device grid.
This module proposes a tensor layout specification for 1D device grids, typically in single-node multi-accelerator configurations, where each complete-information tensor can only be partitioned along a single dimension to prevent information loss.
The complete-information tensors mainly consider the MHA and FFN operations within a Transformer layer, including layer inputs ($\bm{X}_{\text{MHA}}$, $\bm{X}_{\text{FFN}}$), layer parameters ($\bm{W}_{\text{qkv}}$, $\bm{W}_{\text{proj}}$, $\bm{W}_{\text{in}}$, $\bm{W}_{\text{out}}$), and layer outputs ($\bm{O}$, $\bm{Z}$), while excluding intermediate results generated during computation by parallel methods.

According to the architecture of Transformer models, the layer inputs and outputs maintain consistent tensor layouts as $b\times s \times h$. The sharding layouts for each single dimension are as follows: ${b}/{p}\times s\times h$, $b\times {s}/{p}\times h$, and $b\times s\times {h}/{p}$.
The layer parameters primarily focus on weight matrices, the core trainable state of Transformer models. 
In MHA operation, the $h\times h$ matrices of $\bm{Q}$, $\bm{K}$, $\bm{V}$ are concatenated to form parameter $\bm{W}_{\text{qkv}}$ with tensor layout $h\times 3h$, expanding to $h\times 3nh$ when considering $n$ attention heads. 
In FFN operation, the weight matrices $\bm{W}_{\text{in}}$ and $\bm{W}_{\text{out}}$ are designed with dimensions $h\times 4h$ and $4h\times h$, correspondingly. 
These two-dimensional tensors can be sharded either row-wise or column-wise.

We systematically analyze the tensor layouts of parallel methods in terms of sharding dimensions (e.g., $b/s/h$ and row/column) and communication patterns (e.g., AllReduce/AllGather/ReduceScatter). 
For instance, Megatron-LM TP partitions the layer parameters $\bm{W}_{\text{proj}}$,  $\bm{W}_{\text{out}}$ row-wise and $\bm{W}_{\text{qkv}}$, $\bm{W}_{\text{in}}$ column-wise to decrease memory consumption. 
The SP method is integrated into Megatron-LM as the TP+SP paradigm, where an AllGather synchronizes activations along the $s$ dimension before MHA and FFN operations, followed by a ReduceScatter communication.
Table~\ref{tab:layout} concludes the tensor layouts analysis of the parallel methods introduced in the related work. 

Since different parallel methods employ inconsistent tensor layouts, switching between parallel methods can lead to layout mismatches between the output tensor of the preceding operator and the input tensor of the subsequent operator. In such cases, a tensor redistribution operation is required, which introduces additional computational overhead and communication costs.
Accordingly, this module establishes the tensor layout specification for the parallel strategy set to follow, which precisely defines standardized tensor sharding dimensions, as shown in the last row of table~\ref{tab:layout}.

Following the tensor layout specification, we construct a compliant parallel strategy set by extending existing parallel methods.
In tensor processing, most parallel methods adopt sequence-dimension ($s$) sharding for layer inputs/outputs. While ZeRO3 typically employs batch-dimension ($b$) partitioning, it can alternatively shard $s$ through integration with sequence parallel methods.
Regarding layer parameters, parallel methods including Megatron-LM CP, DeepSpeed Ulysses, and Colossal-AI SP maintain intact tensors, but can integrate with ZeRO3 to enable tensor sharding. Notably, ZeRO3 requires row-wise sharding of parameter weight matrices. To satisfy this constraint, we transpose both $\bm{W}_{qkv}$ and $\bm{W}_{in}$, ensuring memory-contiguous layouts across devices for proper AllGather execution.
Adhering to tensor layout specifications, the parallel strategy set $\mathcal{P}$ comprises the following 5 elements (as shown in Figure~\ref{fig:overview}).

- \textit{MegatronTS}: Megatron-LM TP+SP;

- \textit{MegatronCZ}: Megatron-LM CP+DeepSpeed ZeRO3;

- \textit{UlyssesZ}: DeepSpeed Ulysses+DeepSpeed ZeRO3;

- \textit{ColossalZ}: Colossal-AI SP+DeepSpeed ZeRO3;

- \textit{METP}: METP. 

These parallel strategies with identical tensor layouts theoretically enable seamless hot-switching without tensor redistribution. 
However, in practice, the parallel strategies in $\mathcal{P}$ are not switchable due to incompatibilities in implementation, such as programming interfaces and communication mechanisms.
To address it, we construct modular parallel strategy function libraries based on functional programming, encapsulating the MHA and FFN computation processes of each parallel strategy into standardized functions, according to Equations (\ref{eq: qkv}-\ref{eq: ffn}).
Specifically, we instantiate the parametric implementations under parallel strategy $\pi$ for MHA and FFN operations through the mapping: 
\begin{align}
     & f_{\pi, \text{MHA}}:\bm{X}_\text{MHA} \times \bm{W}_\text{MHA} \rightarrow \bm{O}   \\
     & f_{\pi,\text{FFN}}:\bm{X}_\text{FFN} \times \bm{W}_\text{FFN}  \rightarrow \bm{Z}
\end{align}
where $\bm{W}_\text{MHA}  = \{\bm{W}_{\text{qkv}}, \bm{W}_{\text{proj}}\}$,  $\bm{W}_\text{FFN} = \{\bm{W}_{\text{in}}, \bm{W}_{\text{out}}\}$.
For $\forall \pi \in \mathcal{P}$, the functions $f_{\pi,\text{MHA}}$ and $f_{\pi,\text{FFN}}$ are instantiated as the ParaDySeMHA and ParaDySeFFN modules respectively, collectively forming modular function libraries supporting multiple parallel strategies.

\subsection{Hybrid Cost Estimation Module} 

This module proposes sequence-aware hybrid cost models that adaptively select between \textit{Random Forest Regressor} in ensemble learning framework~\cite{Breiman2001} (RF) and \textit{Polynomial Regression} (PR).
The parallel strategy space $\mathcal{P}$ is converted into a multi-dimensional discrete feature via one-hot encoding, and then concatenated with continuous LLM hyperparameters $\mathcal{C}=(h,n,L)$ to form configuration groups $(\mathcal{C},\pi)$. 

For each configuration group, we perform comprehensive profiling to measure the time cost $\mathcal{T}(\bm{X})$ and memory cost $\mathcal{M}(\bm{X})$ on training every $\bm{X} \in \mathcal{X}$ with length $s$ in the dataset. The OOM threshold is set to the maximum memory consumption $OOM(\mathcal{M}_\pi)$. Based on measurements under given configuration group $(\mathcal{C},\pi)$, we independently construct hybrid models for estimating time and memory costs, taking $s$ as the sole input variable.
\begin{equation}\label{eq:cost}
\left (  \begin{matrix} \mathcal{T}(\bm{X})
 \\
\mathcal{M}(\bm{X})
\end{matrix}\right ) =\begin{cases} RF(\bm{X}) \quad s \leq \text{max}(s_\text{profile})
 \\
PR(\bm{X})\quad s > \text{max}(s_\text{profile})
\end{cases}
\end{equation}

RF is employed for \textit{interpolation}, when the input sequence is within the profiling lengths. It combines all tree outputs through mean aggregation to generate regression predictions. 
PR is employed for \textit{extrapolation}, when the input sequence is beyond profiling lengths. Optional polynomial orders, from one to three, preserve the distinctive variation signature of each configuration while avoiding the bias of a global fit. 

This sequence-aware hybrid cost module combines the interpolation advantages of RF with the extrapolation capabilities of PR. By setting OOM threshold $OOM_\pi^\mathcal{C}$ for each configuration group, it incorporates automatic OOM risk detection to inform subsequent heuristic strategy selection.

\subsection{Adaptive Parallel Strategy Switching Module} 

This module proposes a heuristic parallel strategy switching algorithm to select the optimal layer-wise parallel strategy, based on sequence-aware cost models under LLM configuration $\mathcal{C}$ and sequence inputs with length $s$, as detailed in Algorithm \ref{alg:select}. Candidate strategies are filtered based on device memory constraints, with the time-optimal strategy selected from the feasible set. 

\noindent\textbf{Complexity analysis.}
Lines 6-15 yield a worst-case computational complexity of $O(P^2 \times L^2) = O(L^2)$. 
Lines 12-20 yield a worst-case memory complexity of $O(P \times L)+ O(L) = O(L)$, accounting for both the storage of all possible strategy combinations and the final layer-wise strategy selections.

\noindent\textbf{Optimizations.}
Despite the O($L^2$) theoretical complexity, significant constant factors arise from list operations and memory calculations during execution. To address this, we implement multiple optimizations in our heuristic algorithm.

\textit{Parameter Caching}: A dictionary $D$ stores mappings from $(b,s)$ to strategy, reducing time complexity to $O(1)$ for repeated configurations.

\textit{Pre-sorting Pruning} (Line 1): Strategies inferior in both time and memory are eliminated before evaluation.

\textit{Early Termination} (Line 8): If the fastest strategy by time ranking satisfies memory constraints, it's immediately selected (O(L) best-case complexity).

\textit{OOM Short-circuiting} (Lines 16/19): Memory checks terminate early when any strategy fails constraints, minimizing inner-loop iterations.

\textit{Smoothing} (Line 33): If the time cost of previous strategy is estimated slightly higher than the current optimal strategy within ratio $\gamma$, ParaDySe retains the previous strategy to mitigate potential memory overhead from frequent strategy switching.

\subsection{Implementation}

\noindent\textbf{Functional programming libraries.}\label{sec:func_imple}
The modular function library implements parallel strategies within the ParaDySeMHA class and the ParaDySeFFN class, exposing a strategy-agnostic callable $f_{\pi,q}$, such as $f_{\text{METP},\text{MHA}}$. These functional libraries for parallel strategy set follow identical parameter lists (\texttt{input}: layer input, \texttt{state}: layer parameters) and return types (\texttt{result}: layer output). MHA and FFN functions of each parallel strategy take layer inputs and parameters as function arguments and return layer outputs after performing strategy-specific parallel computations. Internally, distributed computation is achieved through collective communication primitives (e.g., AllReduce, AllGather). This design enables parallel strategy hot-switching without reconfiguration.

\noindent\textbf{Cost model implementations.} 
We measure the wall-clock time cost per sequence and real-time peak GPU memory allocation of PyTorch across combinations of parallel strategies $\pi \in \mathcal{P}$ and model configurations $\mathcal{C}$.

The cost model converts parallel strategies into 4-dimensional one-hot encoded features, concatenated with normalized continuous parameters. We instantiate separate RF (n\_estimators=50, max\_depth=10) for each parallel strategy and prediction target (time/memory), implemented via scikit-learn with custom feature importance tracking.

The polynomial regression automatically fits and caches models per configuration using polyfit in numpy, with degree selection via AIC. The prediction engine implements runtime model switching based on sequence length boundaries stored in a configuration database.

\noindent\textbf{Adaptive parallel strategy switching implementations.} 
The ParaDySe framework implementation enables instantaneous parallel strategy switching through a polymorphic function registry design.
ParaDySe maintains a strategy-to-function mapping dictionary, where each key corresponds to a specific operation implementation function (e.g., $f_{\text{MegatronTS},\text{MHA}}$). During execution, the forward method dynamically retrieves the function of target strategy $\pi$, and invokes the function with consistent inferences.
This design guarantees hot-switching across parallel strategies by updating the strategy parameter, which dynamically switches parallel strategies $\pi \in \mathcal{P}$ without reconstructing the computation graph.

\begin{table*}[tbp]
    \caption{Sequence length distributions and maximums of GRCh38 and GitHubCode.}
\centering
\begin{tabular}{ccccccccc}
    \hline
    Length & (0, 4K) & [4K, 8K) & [8K, 16K) & [16K, 32K) & [32K, 64K) & [64K, 128K) & [128K, $+\infty$) & Max\\ \hline
    GitHubCode & 65.7\% & 14.5\% & 9.8\% & 5.1\% & 2.7\% & 1.1\%  & 1.1\% & 309K \\
    GRCh38   & 3.5\%  & 26.4\%  & 28.7\%  & 21.2\%  & 11.9\%  & 5.5\%  & 1.9\% & 624K \\ 
    \hline
\end{tabular}
\label{tab: dataset}
\end{table*}

\begin{table}[tbp]
\caption{LLM configurations.}
\centering
\begin{tabular}{cccccc}
    \hline
    LLM & $h$ & $n$ & $L$ & $L/$node  \\ \hline
    BERT &  1024&  16&  24 &24  \\ 
    LLaMA &  8192&  64&  80 &8  \\
    GPT & 12288& 96& 96  &8  \\
    \hline
\end{tabular}
\label{tab: LLM}
\end{table}

\section{Experiment}

\subsection{Setting} \label{sec: paradyse experiment setting}

\noindent\textbf{Platforms.}
Our experiments are conducted on a computing node equipped with 8 NVIDIA A100-SXM4-80GB GPUs interconnected via NVLink, supported by an AMD EPYC 7473X 24-Core Processor. The software stack comprised PyTorch 2.5.1, CUDA 12.4, NCCL 2.21.5, and FlashAttention-v2.7.4.post1.
We set the random seed to 42 for RF.

\noindent\textbf{Datasets.} 
Table~\ref{tab: dataset} shows that datasets GitHubCode~\cite{codeparrotgithubcode} and GRCh38~\cite{grch38_ncbi} exhibit a pronounced long-tail distribution in length, with max 309K and 624K.
GitHubCode adopts GPT-2 tokenization~\cite{radford2019language}, and GRCh38 adopts 3-mer tokenization~\cite{DNABERT}.
Adopting Curriculum Learning principles~\cite{9392296}, we sort training sequences by length (short-to-long) to enhance training stability.

\noindent\textbf{Models.}
We conduct experiments on three representative LLMs, BERT~\cite{devlin-etal-2019-bert}, LLaMA~\cite{grattafiori2024llama3herdmodels}, GPT~\cite{10.5555/3495724.3495883} as shown in Table~\ref{tab: LLM}.

\noindent\textbf{Baselines.}
We consider individual parallel strategies $\pi \in \mathcal{P}$ as baselines for comparison.
Notably, ColossalZ strategy implements MHA operation by RSA, which inherently conflicts with FlashAttention optimization, resulting in suboptimal quadratic memory complexity. 
Meanwhile, the manually pre-defined switching mechanism of HotSPa inherently restricts support to conventional parallel paradigms and strategies, leaving extremely long genomic or code sequences exceeding 300K tokens beyond its scope. 
Consequently, this section excludes ColossalZ and HotSPa from experimental evaluation and analysis.

\begin{figure*}[t]
    \centering
    \includegraphics[width=\linewidth]{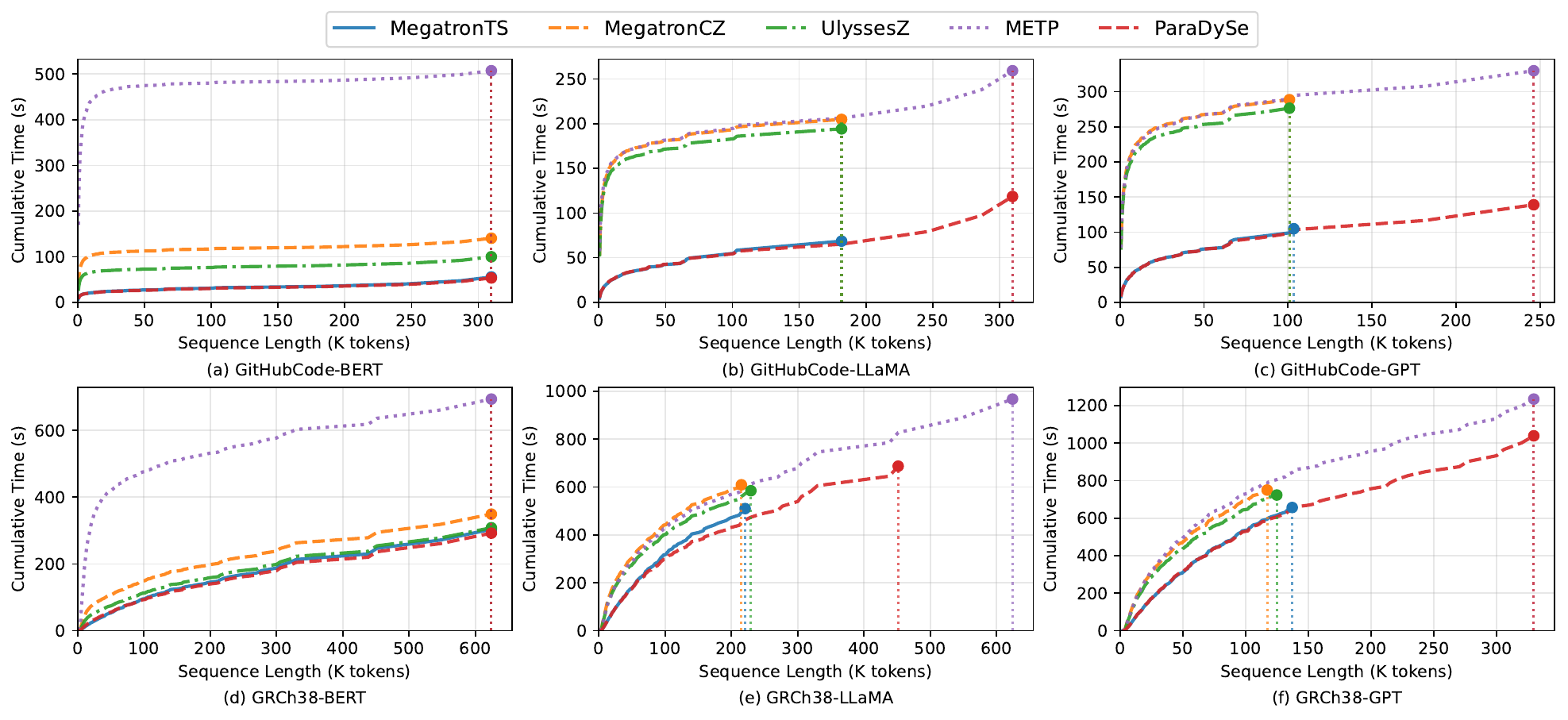}
    \caption{Comparing ParaDySe with baselines across LLMs and long-sequence datasets on cumulative time (s) and sequence length (K tokens). Curve termination indicates OOM failure.} 
    \label{fig:overall}
\end{figure*}

\subsection{Overall Experiment}
Our experiments evaluate the performance and maximum supported sequence length of ParaDySe against baselines by training representative LLMs on datasets with extremely long sequences, as shown in Figure~\ref{fig:overall}.

Experimental results on GitHubCode demonstrate that ParaDySe achieves the longest sequence support among all baselines (resolving OOM failures) while delivering substantial performance improvements (eliminating CPC issues). 
Taking BERT as an example, which is successfully trained by all baselines on the complete dataset, ParaDySe exhibits optimal training efficiency, reducing training time by up to 89.29\% for the maximum sequences compared to baselines. 
As LLM scaling from BERT to GPT, ParaDySe consistently maintains long-sequence support of METP. Notably, with the largest GPT model, ParaDySe extends the trainable sequence length by up to 144.16\% beyond baselines.

ParaDySe demonstrates consistent performance gains on GRCh38, with 58.01\% faster training time for BERT and 181.22\% longer sequences for GPT. ParaDySe effectively enhances training efficiency while substantially expanding sequence length support. We observe a single outlier case during LLaMA training, where the supported sequence length of ParaDySe slightly trails the top-performing baseline, directing future work toward communication analysis.

\subsection{Ablation Study}

\begin{table}[tbp]
\begin{threeparttable}[b]
\caption{Ablation studies on parallel strategy set (L2-L5), hybrid cost model (L6), and algorithm optimization (L7). }
\begin{tabular}{c@{\hspace{0.25cm}}c@{\hspace{0.3cm}}c@{\hspace{0.2cm}}c@{\hspace{0.15cm}}c}  
\hline
Framework        & Seq\_len & Time & Time\_full & Saving \\ \hline
\textbf{ParaDySe (full)}  &    329728   &   3117.76   &     3117.76    & -      \\
w/o MegatronTS   &    300032 &  3173.80 &     2799.45       &  11.80\%      \\
w/o MegatronCZ   &    300032 &   2795.91     &    2799.45       &   -0.13\%     \\
w/o UlyssesZ     &    300032          &  2813.13    &   2799.45           &     0.49\%   \\
w/o METP         &     112128         &  1741.99    &    1724.13  &    1.03\%    \\
w/o RF           &    270336    &  2657.04   &       2622.89       &   1.29\%     \\
w/o Smoothing    &     300032         &    2800.49  &    2799.45  &  0.04\%      \\
\hline
\end{tabular}
\begin{tablenotes}
 \item *Seq\_len: maximum supported sequence length (token); 
 \item *Time: cumulative training time (s) under framework; 
 \item *Time\_full: cumulative training time (s) for ParaDySe (full) to reach the corresponding Seq\_len; 
 \item *Saving: time-saving ratio (\%) of ParaDySe (full) compared to the corresponding framework;
 \item *w/o means without, denoting the ablated component removed from the ParaDySe (full) framework.
\end{tablenotes}
\end{threeparttable}
\label{tab:albation}
\end{table}

The ablation studies comprehensively validate the components of ParaDySe framework. In these studies, we train GPT on GRCh38, and refer to the original ParaDySe framework as ParaDySe (full). 

\noindent\textbf{Parallel strategy set.}
The ablation study confirms that each option in the parallel strategy set contributes significantly to framework performance. Notably, the MegatronTS strategy demonstrates the most substantial impact on computational efficiency, while the METP optimization provides the greatest enhancement to maximum trainable sequence length.

\noindent\textbf{Hybrid cost model.}
The experimental results demonstrate that ParaDySe's hybrid cost model, which integrates RF with PR, achieves superior performance compared to using PR alone. This improvement validates the effectiveness of our combined modeling approach.

\noindent\textbf{Algorithm optimization.}
We set the ratio $\gamma=5\%$ to evaluate smoothing optimization.
Experimental results confirm that the smoothing optimization for strategy switching yields modest but consistent improvements in both model performance and sequence length support. The smoothing optimization contributes to a dramatic enhancement of system stability, effectively eliminating the risks of performance fluctuations and communication overhead caused by abrupt strategy switching during training.
During validation, we identified that frequent strategy switching incurs non-negligible memory overhead and reduces maximum trainable sequence length.

\subsection{Case Study} 

Conventional parallel training frameworks suffer severe inefficiencies when handling OOM failures due to their inability to perform on-the-fly strategy switching, requiring complete environment re-initialization for strategy transitions. Our measurements on GPT training with GRCh38 reveal a total reset latency exceeding 31 seconds, comprising: 2-3 seconds for dependency imports, 5 seconds for dataset reloading, 22 seconds for model reinitialization (accounting for 70.2\% of total overhead), and 1 second for Adam optimizer setup. This rigid switching paradigm creates a critical performance bottleneck.

ParaDySe framework addresses this through an innovative hot-switching mechanism that combines: proactive OOM prediction using trained models to anticipate memory constraints, and switchable functional library with specified tensor layout enabling seamless strategy transitions.
By continuously evaluating memory and computational costs during training, ParaDySe performs smooth hot-switching without interrupting training.

\section{Conclusion}
We propose a novel layer-wise Parallel strategy switching framework for Dynamic Sequences (ParaDySe), which is architecturally composed of three integrated modules.
Hot-Switching Functional Parallelism Module unifies tensor layouts across mainstream parallel methods, enabling layout-compatible strategy switching at Transformer-layer granularity.
Hybrid Cost Estimation Module constructs sequence-aware memory and time cost models to accurately predict resource consumption for each parallel strategy given input lengths.
Adaptive Parallel Strategy Switching Module selects the optimal strategy set for each layer, which balances efficiency and memory utilization, achieving seamless redistribution-free switching.

ParaDySe resolves two critical bottlenecks in LLM training: CPC issues for short sequences and OOM failures for long sequences. Through systematic evaluations of typical LLMs with sequence lengths up to 624K tokens, our framework demonstrates significant improvements in end-to-end training efficiency. By integrating memory-efficient optimizations with general parallel strategies, ParaDySe establishes a new adaptive hot-switching paradigm for efficient and memory-saving LLM scaling.

The memory modeling during experiments was relatively coarse-grained, precluding precise calculations and resulting in suboptimal timing for strategy switching, which led to diminished effectiveness of layer-wise transitions.

\section{Acknowledgments}

This work is sponsored in part by the National Natural Science Foundation of China under Grant No. 62025208 and 62421002.

\bibliography{aaai2026}

\end{document}